\documentclass[11pt,a4paper]{article}
\usepackage[hyperref]{emnlp-ijcnlp-2019}
\usepackage{times}
\usepackage{latexsym}
\usepackage{color}
\usepackage{amsfonts}
\usepackage{graphicx}
\usepackage{amsmath,amsfonts,amssymb,bm}
\usepackage{booktabs}
\usepackage{array}
\usepackage{boxedminipage}
\usepackage{paralist}
\usepackage{url}
\usepackage{diagbox}
\usepackage{enumitem}
\usepackage{paralist}
\definecolor{darkspringgreen}{rgb}{0.05, 0.5, 0.06}
\usepackage{array}
\usepackage{pifont}
\usepackage[caption=false]{subfig}

\usepackage{microtype}      % microtypography
\usepackage{comment}        % bulk comment
\usepackage{amsopn}
\usepackage{xcolor}
\usepackage{multirow}
\usepackage{tikz,pgfplots}

\usepackage{tabularx}
\usepackage{colortbl}
\usepackage{hhline}
\usepackage{url}

\aclfinalcopy % Uncomment this line for the final submission

\newcommand{\sys}{{\sc DyGIE}}
\newcommand{\ours}{{\sc DyGIE++}}
\newcommand{\PreserveBackslash}[1]{\let\temp=\\#1\let\\=\temp}

\newcolumntype{Y}{>{\centering\arraybackslash}X}
\newcolumntype{C}[1]{>{\PreserveBackslash\centering}p{#1}}
\newcolumntype{R}[1]{>{\PreserveBackslash\raggedleft}p{#1}}
\newcolumntype{L}[1]{>{\PreserveBackslash\raggedright}p{#1}}
\newcommand{\bA}{{\mathbf{A}}}

\newcommand{\bd}{{\mathbf{d}}}

\newcommand{\bg}{{\mathbf{g}}}

\newcommand{\bh}{{\mathbf{h}}}

\newcommand{\bu}{{\mathbf{u}}}

\newcommand{\bV}{{\mathbf{V}}}

\newcommand{\bW}{{\mathbf{W}}}

\newcounter{actr}
{\begin{list}{(\alph{actr})}{\usecounter{actr}}}{\end{list}}

\newcounter{ictr}
{\begin{list}{(\roman{ictr})}{\usecounter{ictr}}}{\end{list}}

{\noindent{\em Proof: } \begin{singlespace} \small \noindent}%
{\noindent\qed \end{singlespace}}
\newenvironment{new-proof}[1]%
{{\em Proof of #1: } \begin{singlespace} \small \noindent}%
{\ \noindent\qed \end{singlespace}}

\newcommand{\qed}{\rule[0.1ex]{1.4ex}{1.6ex}}

\newcommand{\reals}{\mathbb{R}}

\title{Entity, Relation, and Event Extraction \\ with Contextualized Span Representations}

\author{
  David Wadden$^\dag$ \quad Ulme Wennberg$^\dag$ \quad Yi Luan$^\ddag$ \quad Hannaneh Hajishirzi$^{\dag*}$ \\
    $^\dag$Paul G. Allen School of Computer Science \& Engineering, University of Washington\\
    $^\ddag$Google AI Language $^{*}$Allen Institute for Artificial Intelligence\\
  \texttt{\{dwadden,ulme,hannaneh\}@cs.washington.edu} \\
  \texttt{luanyi@google.com}\\
}

\date{}

\begin{document}
\maketitle

\begin{abstract}

  We examine the capabilities of a unified, multi-task framework for three information extraction tasks: named entity recognition, relation extraction, and event extraction. Our framework (called \ours) accomplishes all tasks by enumerating, refining, and scoring text spans designed to capture local (within-sentence) and global (cross-sentence) context. Our framework achieves state-of-the-art results across all tasks, on four datasets from a variety of domains. We perform experiments comparing different techniques to construct span representations. Contextualized embeddings like BERT perform well at capturing relationships among entities in the same or adjacent sentences, while dynamic span graph updates model long-range cross-sentence relationships. For instance, propagating span representations via predicted coreference links can enable the model to disambiguate challenging entity mentions. Our code is publicly available at \url{https://github.com/dwadden/dygiepp} and can be easily adapted for new tasks or datasets.

\end{abstract}

\section{Introduction}

Many information extraction tasks  -- including named entity recognition, relation extraction, event extraction, and  coreference resolution -- can benefit from incorporating global context across sentences or from non-local dependencies among phrases. For example, knowledge of a coreference relationship can provide information to help infer the type of a difficult-to-classify entity mention.  In event extraction, knowledge of the entities present in a sentence can provide information that is useful for predicting event triggers.  

To model global context, previous works have used pipelines to extract syntactic, discourse, and other hand-engineered features as inputs to structured prediction models~\cite{Li2013JointEE, yang2016joint, li2014incremental} and neural scoring functions~\cite{Nguyen2019OneFA}, or as a guide for the construction of neural architectures ~\cite{peng2017cross, zhang2018graph, Sha2018JointlyEE, christopoulou2018walk}. Recent end-to-end systems have achieved strong performance by dynmically constructing graphs of spans whose edges correspond to  task-specific relations~\cite{luan2019general, lee2018higher,qian2019graphie}.  

\begin{figure}[t]
  \centering
  \includegraphics[width=\columnwidth, keepaspectratio, trim={0.3cm 3.75cm 4.5cm 5cm}, clip]{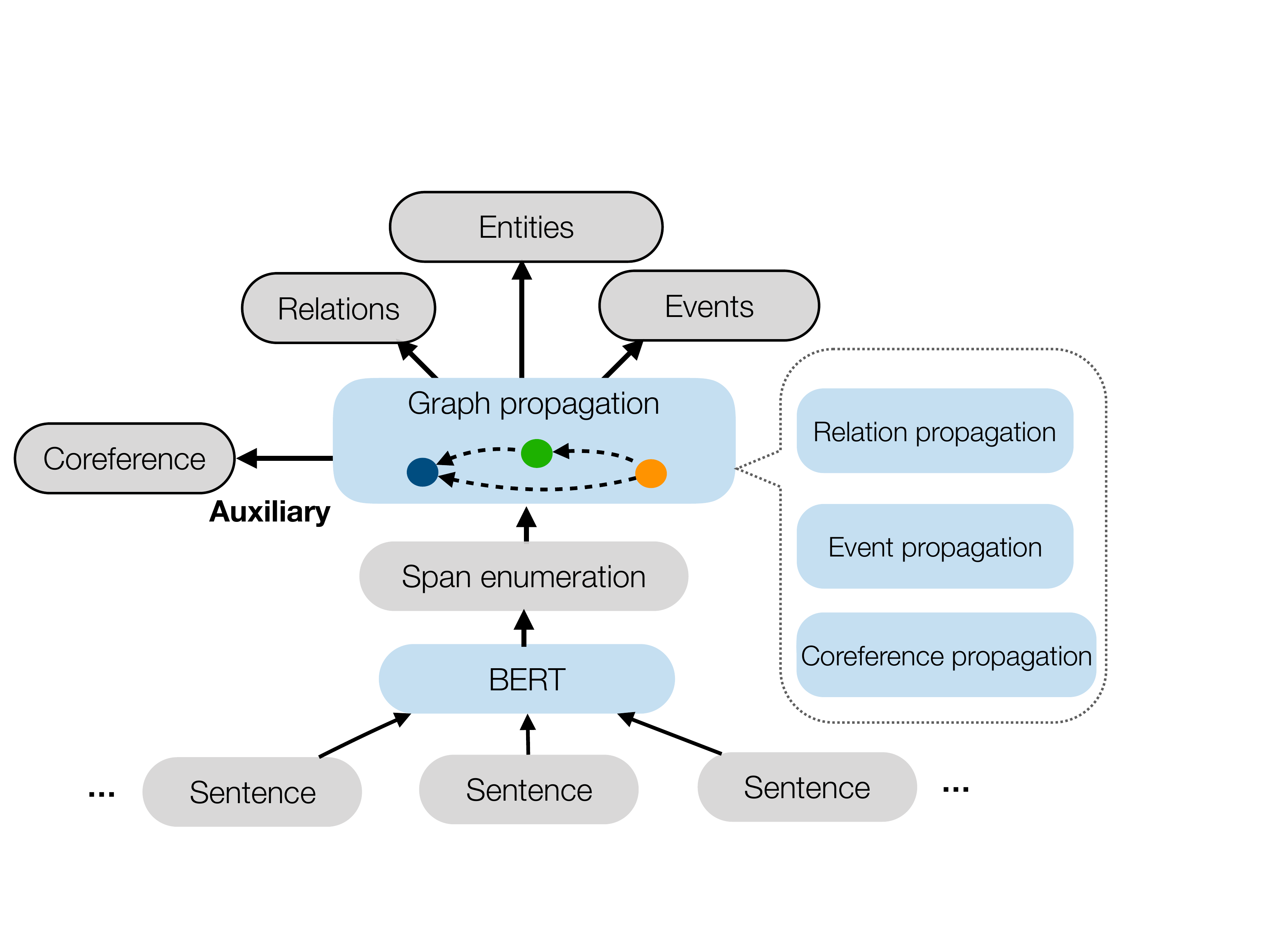}
  \caption{
    \textbf{Overview of our framework}: {\sc \textbf{DyGIE++}}. Shared span representations are constructed by refining contextualized word embeddings via span graph updates, then passed to scoring functions for three IE tasks.
  }
  \label{fig:model}
  \vspace{-1em}
\end{figure}

Meanwhile, contextual language models~\cite{dai2015semi,peters2017semi, peters2018deep, devlin2019bert} have proven successful on a range of natural language processing tasks~\cite{bowman2015large, sang2003introduction, rajpurkar2016squad}. 
Some of these models are also capable of modeling context beyond the sentence boundary. For instance, the attention mechanism in BERT's transformer architecture can capture relationships between tokens in nearby sentences. 

In this paper, we study different methods to incorporate global context in a general multi-task IE framework, building upon a previous span-based IE method~\cite{luan2019general}. Our \ours\ framework, shown in Figure~\ref{fig:model}, enumerates candidate text spans and encodes them using contextual language models and task-specific message updates passed over a text span graph.  Our framework achieves state-of-the results across three IE tasks, leveraging the benefits of both contextualization methods.  

We conduct experiments and a thorough analysis of the model on named entity, relation, and event extraction. Our findings are as follows: (1) Our general span-based framework produces state-of-the-art results on all tasks and all but one subtasks across four text domains, with relative error reductions ranging from 0.2 - 27.9\%. (2) BERT encodings are able to capture important within and adjacent-sentence context, achieving improved performance by increasing the input window size. (3) Contextual encoding through message passing updates enables the model to incorporate cross-sentence dependencies, improving performance beyond that of BERT alone, especially on IE tasks in specialized domains.
\section{Task and Model}
\vspace{-.1cm}
Our \ours framework\ extends a recent span-based model for entity and relation extraction ~\cite{luan2019general} as follows: (1) We perform event extraction as an additional task and propagate span updates across a graph connecting event triggers to their arguments. (2) We build  span representations on top of multi-sentence BERT encodings.
\vspace{-.2cm}

\subsection{Task definitions}

The input is a document represented as a sequence of tokens $D$, from which our model constructs spans $S=\{s_1,\ldots, s_T\}$, the set of all possible within-sentence phrases (up to a threshold length) in the document.

{ Named Entity Recognition} involves predicting the best entity type label $e_i$ for each span $s_i$. For all tasks, the best label may be a ``null'' label.
{Relation Extraction} involves predicting the best relation type $r_{ij}$ for all span pairs $(s_i,s_j)$. For the data sets studied in this work, all relations are between spans within the same sentence. The coreference resolution task is to predict the best coreference antecedent $c_i$ for each span $s_i$. We perform coreference resolution as auxiliary task, to improve the representations available for the ``main'' three tasks.

{ Event Extraction} involves predicting named entities, event triggers,  event arguments, and argument roles.  Specifically, each token $d_i$ is predicted as an event trigger by assigning it a label $t_i$. Then, for each trigger $d_i$,  event arguments are assigned to this event trigger by predicting an argument role $a_{i j}$ for all spans $s_j$ in the same sentence as $d_i$.
Unlike most work on event extraction, we consider the realistic setting where gold entity labels are not available. Instead, we use predicted entity mentions as argument candidates.

\subsection{DyGIE++ Architecture} Figure~\ref{fig:model} depicts the four-stage architecture. For more details, see ~\cite{luan2019general}.
% \begin{itemize}[leftmargin=*, topsep=0pt,noitemsep]

\vspace{.1cm}
\noindent  {\bf Token encoding:} \ours\ uses BERT for  token representations using a ``sliding window'' approach, feeding each sentence to BERT together with a size-$L$ neighborhood of surrounding sentences.

\vspace{.1cm}
\noindent  {\bf Span enumeration:} Spans of text are enumerated and constructed by concatenating the tokens representing their left and right endpoints, together with a learned span width embedding.

\vspace{.1cm}
\noindent{\bf Span graph propagation:} A graph structure is generated dynamically based on the model's current best guess at the relations present among the spans in the document.
Each span representation $\bg_j^t$ is updated by integrating   span representations from its neighbors in the graph according to three variants of graph propagation. In coreference propagation, a span's neighbors in the graph are its likely coreference antecedents.  In relation propagation, neighbors are related entities within a sentence. In event propagation, there are event trigger nodes and event argument nodes; trigger nodes pass messages to their likely arguments, and arguments pass messages back to their probable triggers. The whole procedure is trained end-to-end, with the model learning simultaneously how to identify important links between spans and how to share information between those spans.

More formally,  at each iteration $t$ the model generates an update $\bu_x^t(i)$ for span $s^t \in \reals^d$:
\begin{equation} \label{eq:prop}
  \bu_x^t(i) = \sum_{j \in B_x(i)} V_x^t(i, j) \odot \bg_j^t,
\end{equation}

\vspace{-0.5em}

\noindent where $\odot$ denotes elementwise multiplication and $V_x^t(i, j)$ is a measure of similarity between spans $i$ and $j$ under task $x$ -- for instance, a score indicating the model's confidence that span $j$ is the coreference antecedent of span $i$. For relation extraction, we use a ReLU activation  to enforce sparsity. The final updated span representation $\bg_j^{t+1}$ is computed as a convex combination of the previous representation and the current update, with weights determined by a gating function.

\vspace{.1cm}
\noindent {\bf Multi-task classification:} The re-contextualized representations are input to scoring functions which make predictions for each of the end tasks. We use a two-layer feedforward neural net (FFNN) as the scoring function. For trigger and named entity prediction for span $\bg_i$, we compute $\text{FFNN}_{\text{task}} (\bg_i)$. For relation and argument role prediction, we concatenate the relevant pair of embeddings and compute $\text{FFNN}_{\text{task}}([\bg_i, \bg_j])$.

\vspace{-.1cm}
\section{Experimental Setup}
\paragraph{Data}
We experiment on four different datasets: ACE05, SciERC, GENIA and WLPC (Statistics and details on all data sets and splits can be found in Appendix \ref{sec:data}.). The \textbf{ACE05}  corpus provides entity, relation, and event annotations for a collection of documents from a variety of domains such as newswire and online forums. For named entity and relation extraction we follow the train / dev / test split from \newcite{miwa2016end}. Since the ACE data set lacks coreference annotations, we train on the coreference annotations from the  OntoNotes dataset~\cite{pradhan2012conll}. For event extraction we use the split described in \newcite{yang2016joint, Zhang2019JointEA}. 
We refer to this split as ACE05-E in what follows. The \textbf{SciERC} corpus \cite{luan2018multi} provides entity, coreference and relation annotations from 500 AI paper abstracts. The \textbf{GENIA} corpus~\cite{Kim2003GENIAC} provides entity tags and coreferences for 1999 abstracts from the biomedical research literature with a substantial portion of entities (24\%) overlapping some other entity. The \textbf{WLPC} dataset provides entity, relation, and event annotations for 622 wet lab protocols~\cite{Kulkarni2018AnAC}. Rather than treating event extraction as a separate task, the authors annotate event triggers as an entity type, and event arguments as relations between an event trigger and an argument. 

\vspace{-.2cm}

\paragraph{Evaluation} We follow the experimental setups of the respective state-of-the-art methods for each dataset: \newcite{luan2019general} for entities and relations, and \newcite{Zhang2019JointEA} for event extraction. 
An entity prediction is correct if its label and span matches with a gold entity; a relation is correct if both the span pairs and relation labels match with a gold relation triple. An event trigger is correctly identified if its offsets match a gold trigger. An argument is correctly identified if its offsets and event type match a gold argument. Triggers and arguments are correctly classified if their event types and event roles are also correct, respectively. 

\vspace{-.2cm}

\paragraph{Model Variations} We perform experiments with the following  variants of our model architecture. \textbf{BERT + LSTM} feeds pretrained BERT embeddings to a bi-directional LSTM  layer, and the LSTM parameters are  trained together with task specific layers.  
\textbf{BERT Finetune} uses supervised fine-tuning of BERT on the end-task. 
For each variation, we study the effect of integrating different task-specific message propagation approaches.  
\vspace{-.2cm}

\paragraph{Comparisons} For entity and relation extraction, we compare \ours\ against the \sys~ system it extends. \sys\ is a system based on ELMo~\citep{peters2018deep} that uses dynamic span graphs to propagate global context. 
For event extraction, we compare against the method of \newcite{Zhang2019JointEA}, which is also an ELMo-based approach that relies on inverse reinforcement learning to focus the model on more difficult-to-detect events. 

\begin{table}[t]
  \footnotesize
  \centering

  \begin{tabularx}{\columnwidth}{l l *{3}{Y}}
    \toprule
    Dataset & Task & SOTA  & Ours & $\Delta \% $ \\
    \midrule
    \multirow{2}{*}{ACE05} & Entity & 88.4 & \textbf{88.6} & 1.7 \\
    & Relation & 63.2 & \textbf{63.4}& 0.5  \\
    \cmidrule(lr){1-5}
    \multirow{5}{*}{ACE05-Event*} & Entity & 87.1 & \textbf{90.7} & 27.9 \\
    & Trig-ID & 73.9 & \textbf{76.5} & 9.6 \\
    & Trig-C & 72.0 & \textbf{73.6} & 5.7 \\
    & Arg-ID & \textbf{57.2} & 55.4 & -4.2 \\
    & Arg-C & 52.4 & \textbf{52.5} & 0.2 \\
    \cmidrule(lr){1-5}
    \multirow{2}{*}{SciERC} & Entity & 65.2 & \textbf{67.5} & 6.6 \\
    & Relation & 41.6 & \textbf{48.4} & 11.6 \\
    \cmidrule(lr){1-5}
    \multirow{1}{*}{GENIA} & Entity & 76.2 & \textbf{77.9} & 7.1 \\
    \cmidrule(lr){1-5}
    \multirow{2}{*}{WLPC}  & Entity & 79.5 & \textbf{79.7} & 1.0 \\
    & Relation & 64.1 & \textbf{65.9} & 5.0 \\
    \bottomrule
  \end{tabularx}
  \caption{{\sc \textbf{DyGIE++}} \textbf{achieves state-of-the-art results}. Test set F1 scores of best model, on all tasks and datasets. We define the following notations for events: \textit{Trig}: Trigger, \textit{Arg}: argument, \textit{ID}: Identification,  \textit{C}: Classification. * indicates the use of a 4-model ensemble for trigger detection. See Appendix \ref{sec:implementation} for details. The results of the single model are reported in Table 2 (c). We ran significance tests on a subset of results in Appendix \ref{sec:stats}. All were statistically significant except Arg-C and Arg-ID on ACE05-Event.}

  \label{tab:results_summary}
  \vspace{-2em}

\end{table}

\vspace{-.2cm}

\paragraph{Implementation Details} Our model is implemented using AllenNLP~\cite{Gardner2017AllenNLP}. We use $\text{BERT}_{\text{BASE}}$ for entity and relation extraction tasks and use $\text{BERT}_{\text{LARGE}}$ for event extraction. 
For BERT finetuning, we use BertAdam with the learning rates of $1 \times 10^{-3}$ for the task specific layers, and $5.0 \times 10^{-5}$ for BERT. We use a longer warmup period for BERT than the warmup period for task specific-layers and perform linear decay of the learning rate following the warmup period. Each of the feed-forward neural networks has two hidden layers and ReLU activations and 0.4 dropout. We use 600 hidden units for event extraction and 150 for entity and relation extraction (more  details in Appendix \ref{sec:implementation}).

\begin{table}[t]
  \setlength{\tabcolsep}{.25em}
  \footnotesize
  \centering

    \begin{tabularx}{\columnwidth}{l *{4}{Y}}
      \toprule
      & ACE05 & SciERC & GENIA & WLPC \\
      \midrule
      BERT + LSTM & 85.8 & 69.9 & 78.4 & \textbf{78.9} \\
      \;\; +$\texttt{RelProp}$ & 85.7 & 70.5 & - & 78.7 \\
      \;\; +$\texttt{CorefProp}$& 86.3 & \textbf{72.0} & 78.3 & - \\
      BERT Finetune & 87.3 & 70.5 & 78.3 & 78.5 \\
      \;\; +$\texttt{RelProp}$& 86.7 & 71.1 & - & 78.8 \\
      \;\; +$\texttt{CorefProp}$& \textbf{87.5} & 71.1 & \textbf{79.5} & -\\
      \bottomrule
    \end{tabularx}
    \vspace{-1em}
    \caption{F1 scores on NER.}
    \vspace{1em}

    \label{tab:results_ner}

    \begin{tabularx}{\columnwidth}{l *{3}{Y}}
      \toprule
      & ACE05 & SciERC  & WLPC \\
      \midrule
      BERT + LSTM & 60.6 & 40.3 & 65.1 \\
      \;\; +$\texttt{RelProp}$ & 61.9 & 41.1 & 65.3 \\
      \;\; +$\texttt{CorefProp}$& 59.7 & 42.6 & -\\
      BERT FineTune& \textbf{62.1} & 44.3 & 65.4 \\
      \;\; +$\texttt{RelProp}$& 62.0 & 43.0 & \textbf{65.5} \\
      \;\; +$\texttt{CorefProp}$& 60.0 & \textbf{45.3} & -\\
      \bottomrule
    \end{tabularx}
    \vspace{-1em}
    \caption{F1 scores on Relation.}
    \vspace{1em}

    \label{tab:results_relation}

    \begin{tabularx}{\columnwidth}{l *{4}{Y} }
      \toprule
      & Entity & Trig-C & Arg-ID & Arg-C \\
      \midrule

      BERT + LSTM & 90.5 & 68.9 & \textbf{54.1} & \textbf{51.4} \\
      \;\; +$\texttt{EventProp}$& \textbf{91.0} & 68.4 & 52.5 & 50.3 \\
      BERT FineTune & 89.7 & \textbf{69.7} & 53.0 & 48.8 \\
      \;\; +$\texttt{EventProp}$& 88.7 & 68.2 & 50.4 & 47.2\\
      \bottomrule
    \end{tabularx}
    \vspace{-1em}
    \caption{F1 scores on ACE05-E.}

    \label{tab:results_event}
  \caption{\textbf{Comparison of contextualization methods}. All ablations are performed on the dev set except for ACE05-E, where the precedent in the literature is to ablate on test.}\label{tab:ablations}

  \vspace{-1em}
\end{table}

\section{Results and Analyses}
\vspace{-.2cm}
\paragraph{State-of-the-art Results}
Table~\ref{tab:results_summary} shows test set F1 on the entity, relation and event extraction tasks. Our framework establishes a new state-of-the-art on all three high-level tasks, and on all subtasks except event argument identification. Relative error reductions range from 0.2 - 27.9\% over previous state of the art models.

\vspace{-.2cm}

\paragraph{Benefits of Graph Propagation}

Table \ref{tab:results_ner} shows that Coreference propagation (\texttt{CorefProp}) improves named entity recognition performance across all three domains. The largest gains are on the computer science research abstracts of SciERC, which make frequent use of long-range coreferences, acronyms and abbreviations. \texttt{CorefProp} also improves relation extraction on SciERC.

Relation propagation (\texttt{RelProp}) improves  relation extraction performance over pretrained BERT, but does not improve fine-tuned BERT.  We believe this occurs because all relations are within a single sentence, and thus BERT can be trained to model these relationships well.

Our best event extraction results did not use any propagation techniques (Table \ref{tab:results_event}). We hypothesize that event propagation is not helpful due to the asymmetry of the relationship between triggers and arguments.
Methods to model higher-order interactions among event arguments and triggers represent an interesting direction for future work.

\vspace{-.2cm}

\paragraph{Benefits of Cross-Sentence Context with BERT}
\begin{table}[t]
  \footnotesize
  \centering

  \begin{tabularx}{\columnwidth}{l l *{3}{Y}}
    \toprule
    Task & Variation & 1 & 3 \\

    \midrule
    \multirow{2}{*}{Relation} & BERT+LSTM & 59.3 & \textbf{60.6}\\
    &BERT Finetune &  62.0 & \textbf{62.1}\\
    \midrule
    \multirow{2}{*}{Entity} & BERT+LSTM & 90.0 & \textbf{90.5} \\
    &BERT Finetune &  88.8 & \textbf{89.7}\\
    \midrule
    \multirow{2}{*}{Trigger} & BERT+LSTM & \textbf{69.4} & 68.9 \\
    &BERT Finetune & 68.3 & \textbf{69.7}\\
    \midrule
    \multirow{2}{*}{Arg Class} & BERT+LSTM & 48.6 & \textbf{51.4} \\
    &BERT Finetune &  \textbf{50.0} & 48.8\\
    \bottomrule
  \end{tabularx}
  \caption{\textbf{Effect of BERT cross-sentence context}. F1 score of relation F1 on ACE05 dev set and entity, arg, trigger extraction F1 on ACE05-E test set, as a function of the BERT context window size.}

  \label{tab:sent_length}
  \vspace{-1em}

\end{table}

Table~\ref{tab:sent_length} shows that both variations of our BERT model benefit from wider context windows. Our model achieves the best performance with a 3-sentence window across all relation and event extraction tasks.
\vspace{-.2cm}

\paragraph{Pre-training or Fine Tuning BERT Under Limited Resources}
Table~\ref{tab:ablations} shows that fine-tuning BERT generally performs slightly better than using the pre-trained BERT embeddings combined with a final LSTM layer.\footnote{Pre-trained BERT without a final LSTM layer performed substantially worse than either fine-tuning BERT, or using pre-trained BERT with a final LSTM layer.} Named entity recognition improves by an average of 0.32 F1 across the four datasets tested, and relation extraction improves by an average of 1.0 F1, driven mainly by the performance gains on SciERC. On event extraction, fine-tuning decreases performance by 1.6 F1 on average across tasks. We believe that this is due to the high sensitivity of both BERT finetuning and event extraction to the choice of optimization hyperparameters -- in particular, the trigger detector begins overfitting before the argument detector is finished training.

Pretrained BERT combined with an LSTM layer and graph propagation stores gradients on 15 million parameters, as compared to the 100 million parameters in $\text{BERT}_{\text{BASE}}$.
Since the BERT + LSTM + Propagation approach requires less memory and is less sensitive to the choice of optimization hyperparameters, it may be appealing for non-experts or for researchers working to quickly establish a reasonable baseline under limited resources. It may also be desirable in situations where fine-tuning BERT would be prohibitively slow or memory-intensive, for instance when encoding long documents like scientific articles.
\vspace{-.2cm}

\paragraph{Importance of In-Domain Pretraining}
\begin{table}[t]
  \footnotesize
  \centering

  \begin{tabularx}{\columnwidth}{l *{2}{Y}  *{2}{Y}}
    \toprule
    & \multicolumn{2}{c}{SciERC} & \multicolumn{2}{c}{GENIA} \\
    \cmidrule(lr){2-3} \cmidrule(lr){4-5}

    & Entity & Relation  & Entity \\
    \midrule

    Best BERT & 69.8 & 41.9 & 78.4 \\

    Best SciBERT & \textbf{72.0} & \textbf{45.3} & \textbf{79.5} \\
    \bottomrule
  \end{tabularx}
  \caption{In-domain pre-training:  SciBERT vs. BERT}

  \label{tab:ablate_scibert}
  \vspace{-1em}

\end{table}

We replaced BERT~\cite{devlin2019bert} with SciBERT~\cite{beltagy2019scibert} which is pretrained on a large multi-domain corpus of scientific publications. Table~\ref{tab:ablate_scibert} compares the results of BERT and SciBERT with the best-performing model configurations. SciBERT significantly boosts performance for scientific datasets including SciERC and GENIA.
These results indicate that introducing unlabeled text of similar domains for pre-training can significantly improve performance.

\vspace{-.1cm}
\paragraph{Qualitative Analysis}
To better understand the mechanism by which graph propagation improved performance, we examined all named entities in the SciERC dev set where the prediction made by the $\text{BERT} + \text{LSTM} + \texttt{CorefProp}$ model from Table \ref{tab:results_ner} was different from the $\text{BERT} + \text{LSTM}$ model. We found 44 cases where the \texttt{CorefProp} model corrected an error made by the base model, and 21 cases where it introduced an error. The model without \texttt{CorefProp} was often overly specific in the label it assigned, labeling entities as \emph{Material} or \emph{Method} when it should have given the more general label \emph{Other Scientific Term}. Visualizations of the disagreements between the two model variants can be found in Appendix \ref{sec:visualizations}. Figure \ref{fig:coref_prop_scierc} shows an example where span updates passed along a coreference chain corrected an overly-specific entity identification for the acronym ``CCRs''. We observed similar context sharing via \texttt{CorefProp} in the GENIA data set, and include an example in Appendix \ref{sec:visualizations}.

\begin{figure}[t]
  \centering
  \subfloat[The green span \emph{CCRs} in sentence 2 is updated based on its predicted coreference antecedent.]{\includegraphics[width=1.0\columnwidth, keepaspectratio, clip]{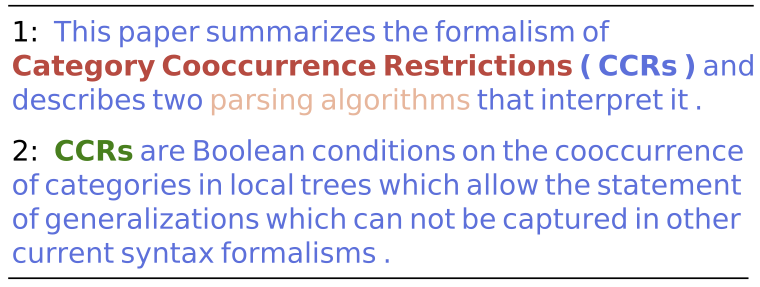}}

  \subfloat[The mention of \emph{CCRs} in sentence 2 serves as a bridge to propagate information from sentence 1 to the mention of \emph{CCRs} in sentence 3]{\includegraphics[width=1.0\columnwidth, keepaspectratio, clip]{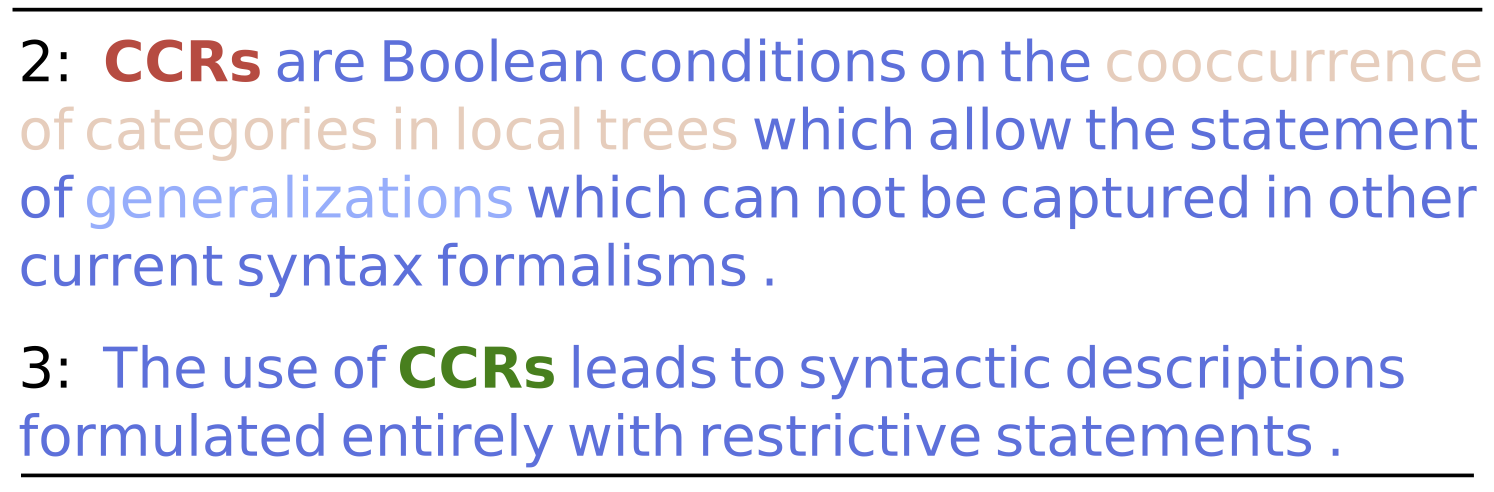}}

  \subfloat[Coreference link strength. Red is strong.]{\includegraphics[width=1.0\columnwidth, keepaspectratio, clip]{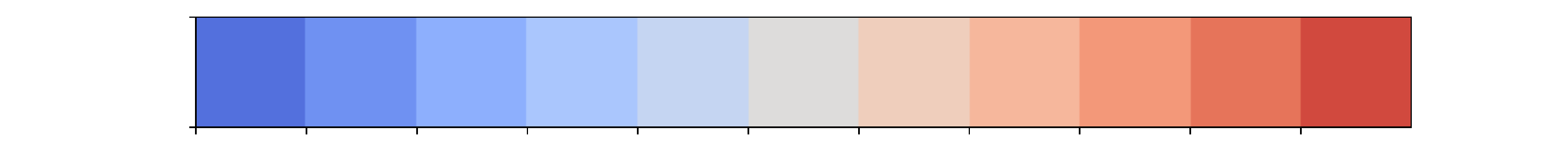}}

  \caption{\textbf{\texttt{CorefProp} enables a correct entity prediction}.
    In each subplot, the green token is being updated by coreference propagation. The preceeding tokens are colored according to the strength of their predicted coreference links with the green token. Tokens in \textbf{bold} are part of a gold coreference cluster discussing \emph{CCRs}. During the \texttt{CorefProp} updates, the span \emph{CCRs} in sentence 2 is updated based on its antecedent \emph{Category Cooccurrence Restrictions}. Then, it passes this information along to the span \emph{CCRs} in sentence 3. As a result, the model changes its prediction for \emph{CCRs} in sentence 3 from \emph{Method} -- which is overly specific according to the \href{http://nlp.cs.washington.edu/sciIE/annotation_guideline.pdf}{SciERC annotation guideline} -- to the correct answer \emph{Other Scientific Term}.
  }

  \label{fig:coref_prop_scierc}
  \vspace{-1em}
\end{figure}

Coreference propagation updated the span representations of all but one of 44 entities, and in 68\% of these cases the update with the largest coreference ``attention weight'' came from a text span in a different sentence that was itself a named entity.

\vspace{-.1cm}
\section{Conclusion}
\vspace{-.1cm}
In this paper, we provide an effective plug-and-play IE framework that can be applied to many information extraction tasks. We explore the abilities of BERT embeddings and graph propagation to capture context relevant for these tasks. We find that combining these two approaches improves performance compared to using either one alone, with BERT building robust multi-sentence representations and graph propagations imposing additional structure relevant to the problem and domain under consideration. Future work could extend the framework to other NLP tasks and explore other approaches to model higher-order interactions like those present in event extraction.

\subsection*{Acknowledgments}
This research was supported by the ONR MURI  N00014-18-1-2670, ONR N00014-18-1-2826, DARPA N66001-19-2-4031, NSF (IIS 1616112), Allen Distinguished Investigator Award, and Samsung GRO. We thank Mandar Joshi for his valuable BERT finetuning advice, Tongtao Zhang for sharing the ACE data code,  anonymous reviewers, and the UW-NLP group for their helpful comments.

\bibliography{references}

\begin{thebibliography}{26}
\expandafter\ifx\csname natexlab\endcsname\relax\def\natexlab#1{#1}\fi

\bibitem[{Beltagy et~al.(2019)Beltagy, Cohan, and Lo}]{beltagy2019scibert}
Iz~Beltagy, Arman Cohan, and Kyle Lo. 2019.
\newblock Scibert: Pretrained contextualized embeddings for scientific text.
\newblock \emph{ArXiv}, abs/1903.10676.

\bibitem[{Bowman et~al.(2015)Bowman, Angeli, Potts, and
  Manning}]{bowman2015large}
Samuel~R. Bowman, Gabor Angeli, Christopher Potts, and Christopher~D. Manning.
  2015.
\newblock A large annotated corpus for learning natural language inference.
\newblock In \emph{EMNLP}.

\bibitem[{Christopoulou et~al.(2018)Christopoulou, Miwa, and
  Ananiadou}]{christopoulou2018walk}
Fenia Christopoulou, Makoto Miwa, and Sophia Ananiadou. 2018.
\newblock A walk-based model on entity graphs for relation extraction.
\newblock In \emph{ACL}.

\bibitem[{Dai and Le(2015)}]{dai2015semi}
Andrew~M Dai and Quoc~V Le. 2015.
\newblock Semi-supervised sequence learning.
\newblock In \emph{NeurIPs}.

\bibitem[{Devlin et~al.(2018)Devlin, Chang, Lee, and
  Toutanova}]{devlin2019bert}
Jacob Devlin, Ming-Wei Chang, Kenton Lee, and Kristina Toutanova. 2018.
\newblock Bert: Pre-training of deep bidirectional transformers for language
  understanding.
\newblock In \emph{NAACL-HLT}.

\bibitem[{Gardner et~al.(2017)Gardner, Grus, Neumann, Tafjord, Dasigi, Liu,
  Peters, Schmitz, and Zettlemoyer}]{Gardner2017AllenNLP}
Matt Gardner, Joel Grus, Mark Neumann, Oyvind Tafjord, Pradeep Dasigi,
  Nelson~F. Liu, Matthew Peters, Michael Schmitz, and Luke~S. Zettlemoyer.
  2017.
\newblock \href {http://arxiv.org/abs/arXiv:1803.07640} {Allennlp: A deep
  semantic natural language processing platform}.

\bibitem[{Kim et~al.(2003)Kim, Ohta, Tateisi, and Tsujii}]{Kim2003GENIAC}
Jin-Dong Kim, Tomoko Ohta, Yuka Tateisi, and Jun'ichi Tsujii. 2003.
\newblock Genia corpus - a semantically annotated corpus for bio-textmining.
\newblock \emph{Bioinformatics}, 19 Suppl 1:i180--2.

\bibitem[{Kulkarni et~al.(2018)Kulkarni, Xu, Ritter, and
  Machiraju}]{Kulkarni2018AnAC}
Chaitanya Kulkarni, Wei Xu, Alan Ritter, and Raghu Machiraju. 2018.
\newblock An annotated corpus for machine reading of instructions in wet lab
  protocols.
\newblock In \emph{NAACL-HLT}.

\bibitem[{Lee et~al.(2018)Lee, He, and Zettlemoyer}]{lee2018higher}
Kenton Lee, Luheng He, and Luke~S. Zettlemoyer. 2018.
\newblock Higher-order coreference resolution with coarse-to-fine inference.
\newblock In \emph{NAACL-HLT}.

\bibitem[{Li and Ji(2014)}]{li2014incremental}
Qi~Li and Heng Ji. 2014.
\newblock Incremental joint extraction of entity mentions and relations.
\newblock In \emph{ACL}.

\bibitem[{Li et~al.(2013)Li, Ji, and Huang}]{Li2013JointEE}
Qi~Li, Heng Ji, and Liang Huang. 2013.
\newblock Joint event extraction via structured prediction with global
  features.
\newblock In \emph{ACL}.

\bibitem[{Luan et~al.(2018)Luan, He, Ostendorf, and Hajishirzi}]{luan2018multi}
Yi~Luan, Luheng He, Mari Ostendorf, and Hannaneh Hajishirzi. 2018.
\newblock Multi-task identification of entities, relations, and coreference for
  scientific knowledge graph construction.
\newblock In \emph{EMNLP}.

\bibitem[{Luan et~al.(2019)Luan, Wadden, He, Shah, Ostendorf, and
  Hajishirzi}]{luan2019general}
Yi~Luan, Dave Wadden, Luheng He, Amy Shah, Mari Ostendorf, and Hannaneh
  Hajishirzi. 2019.
\newblock A general framework for information extraction using dynamic span
  graphs.
\newblock In \emph{NAACL-HLT}.

\bibitem[{Miwa and Bansal(2016)}]{miwa2016end}
Makoto Miwa and Mohit Bansal. 2016.
\newblock End-to-end relation extraction using lstms on sequences and tree
  structures.
\newblock In \emph{ACL}.

\bibitem[{Nguyen and Nguyen(2019)}]{Nguyen2019OneFA}
Trung~Minh Nguyen and Thien~Huu Nguyen. 2019.
\newblock One for all: Neural joint modeling of entities and events.
\newblock In \emph{AAAI}.

\bibitem[{Peng et~al.(2017)Peng, Poon, Quirk, Toutanova, and tau
  Yih}]{peng2017cross}
Nanyun Peng, Hoifung Poon, Chris Quirk, Kristina Toutanova, and Wen tau Yih.
  2017.
\newblock Cross-sentence n-ary relation extraction with graph lstms.
\newblock \emph{Transactions of the Association for Computational Linguistics},
  5:101--115.

\bibitem[{Peters et~al.(2017)Peters, Ammar, Bhagavatula, and
  Power}]{peters2017semi}
Matthew~E. Peters, Waleed Ammar, Chandra Bhagavatula, and Russell Power. 2017.
\newblock Semi-supervised sequence tagging with bidirectional language models.
\newblock In \emph{ACL}.

\bibitem[{Peters et~al.(2018)Peters, Neumann, Iyyer, Gardner, Clark, Lee, and
  Zettlemoyer}]{peters2018deep}
Matthew~E. Peters, Mark Neumann, Mohit Iyyer, Matt Gardner, Christopher Clark,
  Kenton Lee, and Luke Zettlemoyer. 2018.
\newblock Deep contextualized word representations.
\newblock In \emph{NAACL}.

\bibitem[{Pradhan et~al.(2012)Pradhan, Moschitti, Xue, Uryupina, and
  Zhang}]{pradhan2012conll}
Sameer Pradhan, Alessandro Moschitti, Nianwen Xue, Olga Uryupina, and Yuchen
  Zhang. 2012.
\newblock Conll-2012 shared task: Modeling multilingual unrestricted
  coreference in ontonotes.
\newblock In \emph{Joint Conference on EMNLP and CoNLL-Shared Task}, pages
  1--40. Association for Computational Linguistics.

\bibitem[{Qian et~al.(2018)Qian, Santus, Jin, Guo, and
  Barzilay}]{qian2019graphie}
Yujie Qian, Enrico Santus, Zhijing Jin, Jiang Guo, and Regina Barzilay. 2018.
\newblock Graphie: A graph-based framework for information extraction.
\newblock In \emph{NAACL-HLT}.

\bibitem[{Rajpurkar et~al.(2016)Rajpurkar, Zhang, Lopyrev, and
  Liang}]{rajpurkar2016squad}
Pranav Rajpurkar, Jian Zhang, Konstantin Lopyrev, and Percy Liang. 2016.
\newblock Squad: 100, 000+ questions for machine comprehension of text.
\newblock In \emph{EMNLP}.

\bibitem[{Sang and De~Meulder(2003)}]{sang2003introduction}
Erik~F Sang and Fien De~Meulder. 2003.
\newblock Introduction to the conll-2003 shared task: Language-independent
  named entity recognition.
\newblock In \emph{NAACL}.

\bibitem[{Sha et~al.(2018)Sha, Qian, Chang, and Sui}]{Sha2018JointlyEE}
Lei Sha, Feng Qian, Baobao Chang, and Zhifang Sui. 2018.
\newblock Jointly extracting event triggers and arguments by dependency-bridge
  rnn and tensor-based argument interaction.
\newblock In \emph{AAAI}.

\bibitem[{Yang and Mitchell(2016)}]{yang2016joint}
Bishan Yang and Tom~M. Mitchell. 2016.
\newblock Joint extraction of events and entities within a document context.
\newblock In \emph{HLT-NAACL}.

\bibitem[{Zhang et~al.(2019)Zhang, Ji, and Sil}]{Zhang2019JointEA}
Tongtao Zhang, Heng Ji, and Avirup Sil. 2019.
\newblock Joint entity and event extraction with generative adversarial
  imitation learning.
\newblock \emph{Data Intelligence}, 1:99--120.

\bibitem[{Zhang et~al.(2018)Zhang, Qi, and Manning}]{zhang2018graph}
Yuhao Zhang, Peng Qi, and Christopher~D. Manning. 2018.
\newblock Graph convolution over pruned dependency trees improves relation
  extraction.
\newblock In \emph{EMNLP}.

\end{thebibliography}
\bibliographystyle{acl_natbib}
\newpage
\newpage
\clearpage

\appendix

\section{Data} \label{sec:data}

\subsection{Dataset statistics}

Table \ref{tab:data_joint} provides summary statistics for all datasets used in the paper.

\begin{table}[h]
  \centering
  \footnotesize
  \begin{tabular}{L{1.5cm}L{1.0cm}R{0.5cm}R{0.5cm}R{0.5cm}C{0.5cm}C{0.5cm}}
    \toprule
    & Domain & Docs & Ent & Rel & Trig & Arg\\
    \midrule
    ACE05 & News & 511 & 7 & 6 & - & -\\
    ACE05-E & News & 599 & 7 & - & 33 & 22\\
    SciERC & AI & 500 & 6 & 7 & - & -\\
    GENIA & Biomed & 1999 & 5 & - & - & - \\
    WLP & Bio lab & 622 & 18 & 13 & - & -\\
    \bottomrule
  \end{tabular}
  \caption{Datasets for joint entity and relation extraction and their statistics. \emph{Ent}: Number of entity categories. \emph{Rel}: Number of relation categories. \emph{Trig}: Number of event trigger categories. \emph{Arg}: Number of event argument categories.}
  \label{tab:data_joint}
  \vspace{-1em}
\end{table}

\subsection{ACE event data preprocessing and evaluation}

There is some inconsistency in the ACE event literature on how to handle ``time'' and ``value'' event arguments, which are not technically named entities. Some authors, for instance \newcite{yang2016joint} leave them in and create new entity types for them. We follow the preprocessing of \newcite{Zhang2019JointEA}, who ignore these arguments entirely, since these authors shared their preprocessing code with us and report the current state of the art. We will be releasing code at \url{https://github.com/dwadden/dygiepp} to exactly reproduce our data preprocessing, so that other authors can compare their approaches on our data. Due to this discrepancy in the literature, however, our results for named entity and event argument classification are not directly comparable with some previous works.

In addition, there is some confusion on what constitutes an "Event argument identification". Following \newcite{yang2016joint} and \newcite{Zhang2019JointEA}, we say that an argument is identified correctly if its offsets and event type are correct. Some other works seem to require require only that an argument's offsets be identified, not its event type. We do not compare against these.

\section{Graph Propagation} \label{sec:event_graph}

\newcommand{\trig}{\bh_i}
\newcommand{\argu}{\bg_j}
\newcommand{\simil}{\bV_A^t(i, j)}
\newcommand{\similprime}{\bV_A^{t'}(i, j)}
\newcommand{\updateArgTrig}{\bu^t_{A \to T}(i)}
\newcommand{\gateArgTrig}{f_{A \to T}^t(i)}
\newcommand{\weightArgTrig}{\bW_{A \to T}}
\newcommand{\updateTrigArg}{\bu^t_{T \to A}(j)}
\newcommand{\gateTrigArg}{f_{T \to A}^t(j)}

We model relation and coreference interactions similarly to \citet{luan2019general}, and extend the approach to incorporate events. We detail the event propagation procedure here. While the relation and coreference span graphs consist of a single type of node, the event graph consists of two types of nodes: triggers and arguments.

The intuition behind the event graph is to provide each trigger with information about its potential arguments, and each potential argument with information about triggers for the events in which it might participate.\footnote{Event propagation is a somewhat different idea from \cite{Sha2018JointlyEE}, who model argument-argument interactions using a learned tensor. We experimented with adding a similar tensor to our architecture, but did not see any clear performance improvements.}

The model iterates between updating the triggers based on the representations of their likely arguments, and updating the arguments based on the representations of their likely triggers. More formally, denote the number of possible semantic roles played by an event argument (i.e. the number of argument labels) as $L_A$,  $B_T$ as a beam of candidate trigger tokens, and  $B_A$ as a beam of candidate argument spans. These beams are selected by learned scoring functions.
For each trigger $\trig^t \in B_T$ and argument $\argu^t \in B_A$, the model computes a similarity vector $\simil$ by concatenating the trigger and argument embeddings and running them through a feedforward neural network.
The $k^{th}$ element of $\simil$ scores how likely it is that argument $\argu$ plays role $k$ in the event triggered by $\trig$.

Extending Equation~\ref{eq:prop}, the model  updates the trigger $\trig$ by taking an average of the candidate argument embeddings, weighted by the likelihood that each candidate plays a role in the event:
\begin{equation} \label{eqn:trigger_update}
  \updateArgTrig = \sum_{j \in B_A} \bA_{A \to T} f(\simil) \odot \argu^t,
\end{equation}
where $\bA_{A \to T} \in \reals^{d \times L_A}$ is a learned projection matrix, $f$ is a ReLU function, $\odot$ is an elementwise product, and $d$ is the dimension of the span embeddings. Then, the model computes a gate determining how much of the update from (\ref{eqn:trigger_update}) to apply to the trigger embedding:
\begin{equation} \label{eqn:gate}
  \gateArgTrig = \sigma \left(\weightArgTrig [\trig^t, \updateArgTrig] \right),
\end{equation}
where $\weightArgTrig \in \reals^{d \times 2d}$ is a learned projection matrix and $\sigma$ is the logistic sigmoid function. Finally, the updated trigger embeddings are computed as follows:
\begin{equation} \label{eqn:new_span}
  \trig^{t+1} = \gateArgTrig \odot \trig^t + (1 - \gateArgTrig) \odot  \updateArgTrig.
\end{equation}

Similarly, an update for argument span $\argu$ is computed via messages $\updateTrigArg$ as a weighted average over the trigger spans. The update is computed analgously to Equation~\ref{eqn:trigger_update}, with a new trainable matrix $\bA_{T \to A}$.
Finally, the gate $\gateTrigArg$ and the updated argument spans $\argu^{t+1}$ are computed in the same fashion as (\ref{eqn:gate}) and (\ref{eqn:new_span}) respectively. This process represents one iteration of event graph propagation. The outputs of the graph propagation are contextualized trigger and argument representations. When event propagation is performed, the final trigger scorer takes the contextualized surrogate spans $\trig$ as input rather than the original token embeddings $\bd_i$.

\section{\texttt{CorefProp} visualizations} \label{sec:visualizations}

\begin{figure}[t]
  \centering
  \subfloat[\texttt{CorefProp} corrects a mistake.] {
    \includegraphics[width=1.0\columnwidth, keepaspectratio, clip]{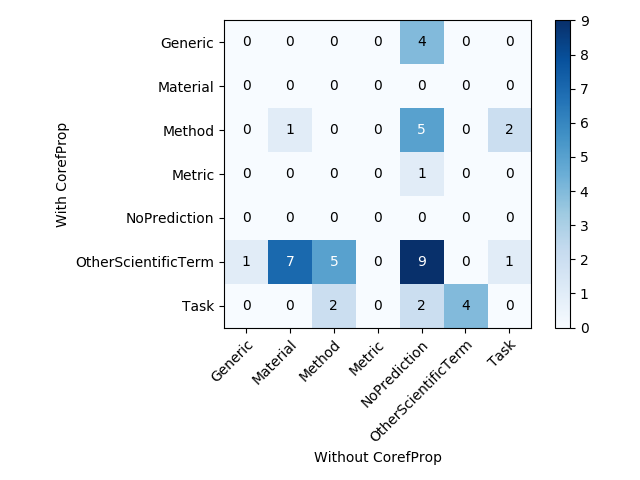}
    \label{fig:bad_to_good}
  }

  \subfloat[\texttt{CorefProp} makes a mistake.] {
    \includegraphics[width=1.0\columnwidth, keepaspectratio, clip]{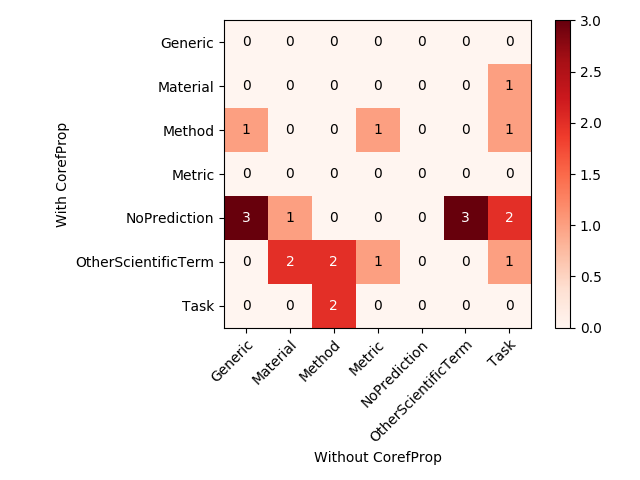}
    \label{fig:good_to_bad}
  }

  \caption{
    Confusion matrix of cases in the SciERC dev set where coreference propagation changed a prediction from incorrect to correct (Fig. \ref{fig:bad_to_good}), or correct to incorrect (Fig. \ref{fig:good_to_bad}). \texttt{CorefProp} leads to more corrections than mistakes, and tends to make less specific predictions (i.e. \emph{OtherScientificTerm}).
  }
  \label{fig:coref_prop_confusion}
\end{figure}

Figure \ref{fig:coref_prop_confusion} shows confusion matrices for cases where \texttt{CorefProp} corrects and introduces a mistake on SciERC named entity recognition. It tends to correct mistakes where the base model either missed an entity, or was overly specific -- classifying an entity as a \emph{Material} or \emph{Method} when it should have been classified with the more general label \emph{OtherScientificTerm}. Similarly, \texttt{CorefProp} introduces mistakes by assigning labels that are too general, or by missing predictions.

\begin{figure}[t]
  \centering
  \subfloat[To classify the mention of \emph{v-erbA} in Sentence 4, the model can share information from its coreference antecedents (in red).] {
    \includegraphics[width=\columnwidth, keepaspectratio, clip]{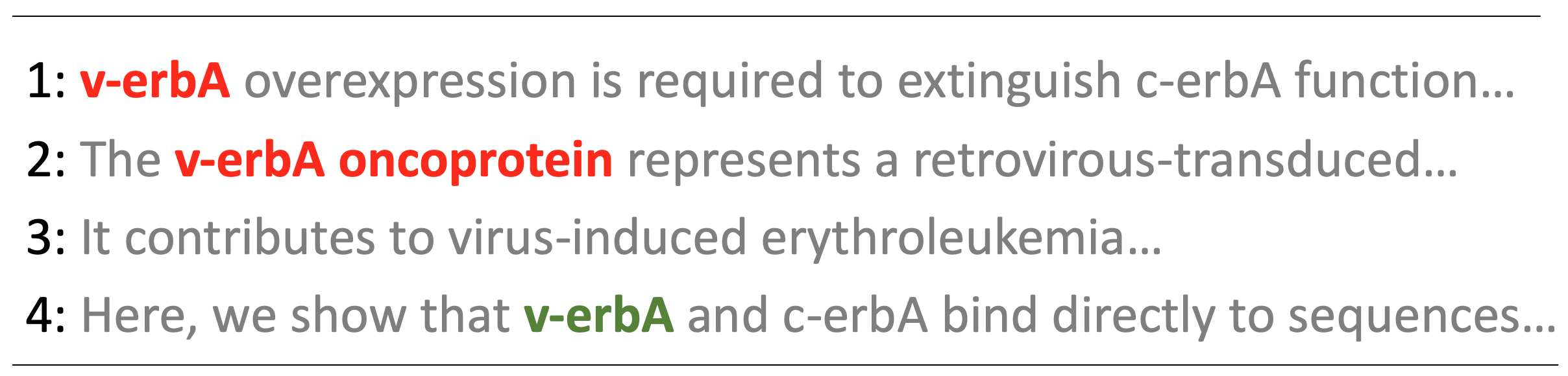}}

  \subfloat[Coreference attention weights. Darker is larger. The biggest update comes from \emph{The v-erbA oncoprotein}] {
    \includegraphics[width=0.9\columnwidth, keepaspectratio, clip]{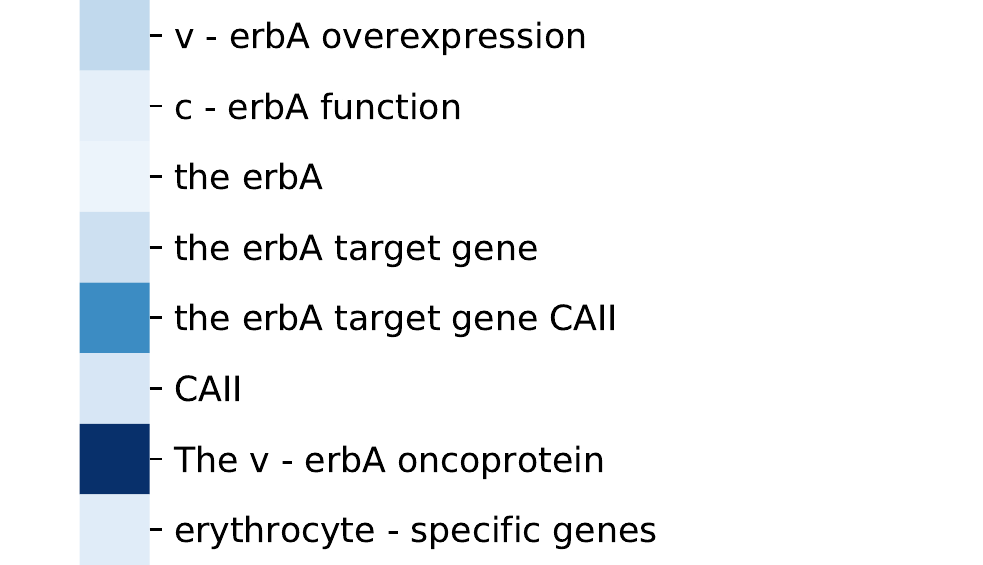}}
  \caption{\textbf{\texttt{CorefProp} aggregates information from informative text spans}.
    By using the representation of the span \emph{v-erbA oncoprotein} in Sentence 2 to update the representation of \emph{v-erbA} in sentence 4, the model is able to correctly classify the latter entity mention as a \emph{protein}.
  }
  \label{fig:coref_attention_weights}
  \vspace{-1em}
\end{figure}

Figure \ref{fig:coref_attention_weights} shows a visualization of the coreference attention weights for a named entity in the GENIA data set. The acronym ``v-erbA'' is correctly identified as a protein, due to a span update from its coreference antecedent ``v-erbA oncoprotein''.

\section{Statistical significance of results} \label{sec:stats}

For a subset of the results in Table \ref{tab:results_summary}, we evaluated statistical significance by re-training a model with 5 random seeds and computing the mean and standard error of the mean of the F1 scores. For ensemble models (trigger detection), we trained 3 ensembles instead of 5 due to the large computational demands of training ensemble models. Due to the large number of experiments performed, it was impractical to perform these tests for every experiment. We report means and standard errors in Table \ref{tab:stats}. For most results, our mean is more than two standard errors above the previous state of the art; those results are significant. Our event argument results are  not significant. For trigger classification, our mean F1 is a little less than two standard errors above the state of the art, indicating moderate significance.

\begin{table}[t]
  \footnotesize
  \centering

  \begin{tabularx}{\columnwidth}{l l *{3}{Y}}
    \toprule
    Dataset & Task & SOTA  & Ours (mean) & Ours (sem) \\
    \midrule
    \multirow{5}{*}{ACE05-Event*} & Entity & 87.1 & 90.4 & 0.1 \\
    & Trig-ID & 73.9 & 76.1 & 0.4 \\
    & Trig-C & 72.0 & 73.0 & 0.6 \\
    & Arg-ID & 57.2 & 54.0 & 0.4 \\
    & Arg-C & 52.4 & 51.3 & 0.4 \\
    \cmidrule(lr){1-5}
    \multirow{2}{*}{SciERC} & Entity & 65.2 & \ 66.3 & 0.4 \\
    & Relation & 41.6 & 46.2 & 0.4 \\
    \bottomrule
  \end{tabularx}
  \caption{Mean and standard error of the mean. Trig-C is moderately significant. Arg-ID and Arg-C do not improve SOTA when averaging across five models. The remaining results are highly significant. Note that our means here differ from the numbers in Table \ref{tab:results_summary}, where we report our best single run to be consistent with previous literature.}

  \label{tab:stats}
  \vspace{-1em}

\end{table}

\section{Implementation Details} \label{sec:implementation}
\paragraph{Learning rate schedules}
For BERT finetuning, we used BertAdam with a maximum learning rate of $1 \times 10^{-3}$ for the task specific layers and $5\times 10^{-5}$ for BERT. For the learning rate schedule, we had an initial warmup period of 40000 batches for the BERT parameters, and 20000 batches for the task specific layers. Following the warmup period, we linearly decayed the learning rate.

For event extraction models with no finetuning we found that SGD with momentum performed better than Adam. We used the PyTorch implementation of SGD, with an initial learning rate of 0.02, momentum of 0.9, weight decay of $1 \times 10^{-6}$ and a batch size of 15 sentences. We cut the learning rate in half whenever dev set F1 had not decreased for 3 epochs.

For all models, we used early stopping based on dev set loss.

\paragraph{Hyperparameter selection}
We experimented with both $\text{BERT}_{\text{BASE}}$ and $\text{BERT}_{\text{LARGE}}$ on all tasks. We found that $\text{BERT}_{\text{LARGE}}$ provided improvement on event extraction with a final LSTM layer, but not on any of the other tasks or event extraction with BERT fine-tuning. In our final experiments we used $\text{BERT}_{\text{BASE}}$ except in the one setting mentioned were $\text{BERT}_{\text{LARGE}}$ was better.

We experiment with hidden layer sizes of 150, 300, and 600 for our feedforward scoring functions. We found that 150 worked well for all tasks except event extraction, where 600 hidden units were used.

\paragraph{Event extraction modeling details} For event extraction we experimented with a final ``decoding'' module to ensure that event argument assignments were consistent with the types of the events in which they participated -- for instance, an entity participating in a ``Personnel.Nominate'' event can play the role of ``Agent'', but not the role of ``Prosecutor''. However, we found in practice that the model learned which roles were compatible with each event type, and constrained decoding did not improve performance. For argument classification, we included the entity label of each candidate argument as an additional feature. At train time we used gold entity labels and at inference time we used the  softmax scores for each entity class as predicted by the named entity recognition model.

\paragraph{Event model ensembling} For the event extraction experiments summarized in Table \ref{tab:results_event} we performed early stopping based on dev set argument role classification performance. However, our trigger detector tended to overfit before the argument classifier had finished training. We also found stopping based on dev set error to be unreliable, due to the small size and domain shift between dev and test split on the ACE05-E data set. Therefore, for our final predictions reported in Table \ref{tab:results_summary}, we trained a four-model ensemble optimized for trigger detections rather than event argument classification, and combined the trigger predictions from this model with the argument role predictions from our original model. This combination improves both trigger detection \emph{and} argument classification, since an argument classification is only correct if the trigger to which it refers is also classified correctly.

\end{document}